\pdfoutput=1

\documentclass[11pt]{article}

\usepackage[final]{acl}

\usepackage{times}
\usepackage{latexsym}

\usepackage[T1]{fontenc}

\usepackage[utf8]{inputenc}

\usepackage{microtype}

\usepackage{inconsolata}

\usepackage{graphicx}
\usepackage{array}
\usepackage{geometry}
\usepackage{booktabs}
\usepackage{amsmath}
\usepackage{pifont}
\usepackage{arydshln}
\usepackage{tabularx}
\usepackage{adjustbox}
\usepackage{etoolbox}
\usepackage{tikz}
\usepackage{mdframed}
\usepackage{lipsum}
\usepackage{multirow}
\usepackage{enumitem}
\usepackage{enumitem}
\usepackage{makecell}
\usepackage{float}
\usepackage[]{hyperref}

\newcolumntype{P}[1]{>{\centering\arraybackslash}p{#1}}

\usetikzlibrary{shapes,arrows}

\usetikzlibrary{shapes.geometric, arrows, positioning}

\newcolumntype{H}{>{\setbox0=\hbox\bgroup}c<{\egroup}@{}}

\newmdtheoremenv{theo}{Theorem}
\usepackage{tabularx}

\usepackage[most]{tcolorbox}

\newtcolorbox[auto counter,number within=subsection]{myBox}[3][]{
arc=2.5mm,
lower separated=false,
fonttitle=\bfseries,
colbacktitle=white!10,
coltitle=black!20!black,
enhanced,
attach boxed title to top left={xshift=0.3cm,
        yshift=-2mm},
colframe=black!20!black,
colback=white
}

\tcbset{
    keytakeaways/.style={
        colback=blue!1, 
        colframe=blue!30, 
        fonttitle=\bfseries, 
        title=Key Takeaways, 
        boxrule=0.5mm, 
        arc=4mm, 
        left=1mm, 
        right=1mm, 
        top=1mm, 
        bottom=1mm, 
        coltitle=black 
    }
}

\usepackage{ifthen}
 \newboolean{showcomments}
 \setboolean{showcomments}{true} 
\newcommand{\harsh}[1]{\ifthenelse{\boolean{showcomments}}{\textcolor{green}{[#1 —harsh]}}{}}

\newcommand{\raj}[1]{\ifthenelse{\boolean{showcomments}}{\textcolor{purple}{[#1 —raj]}}{}}

\newcommand{\aashish}[1]{\ifthenelse{\boolean{showcomments}}{\textcolor{cyan}{[#1 —aashish]}}{}}

\newcommand{\newtext}[1]{\ifthenelse{\boolean{showcomments}}{\textcolor{cyan}{[#1 —newtext]}}{}}

\title{From Intentions to Techniques: A Comprehensive Taxonomy and Challenges in Text Watermarking for Large Language Models}

\newcommand{\gtlogo}{\raisebox{3.4pt}{\includegraphics[scale=0.04]{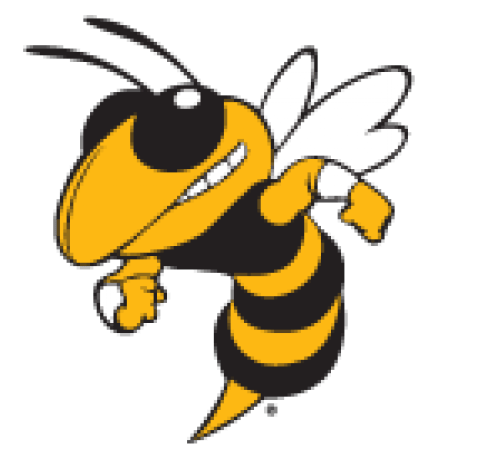}}}
\newcommand{\psulogo}{\raisebox{3.4pt}{\includegraphics[scale=0.025]{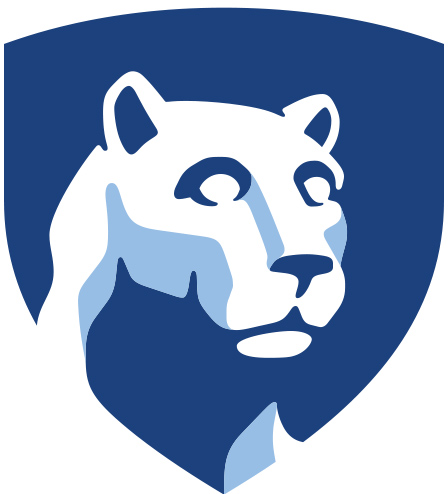}}}
\newcommand{\bitslogo}{\raisebox{3.4pt}{\includegraphics[scale=0.02]{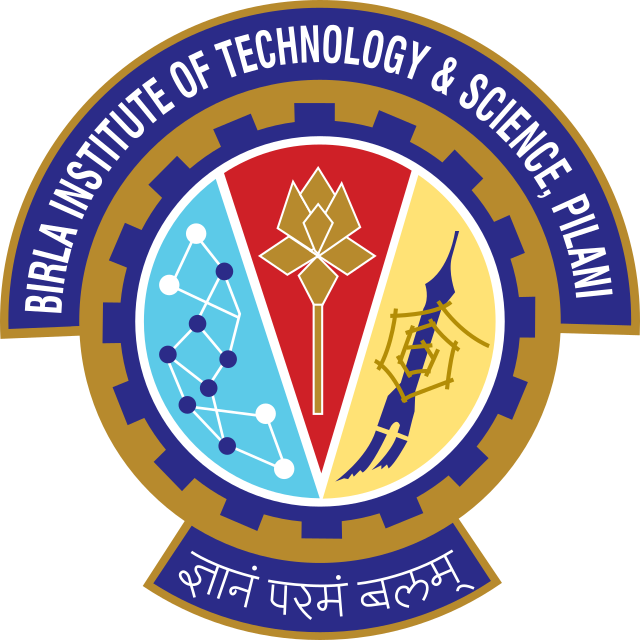}}}

\definecolor{myColour}{HTML}{003366}

\makeatletter
\def\thanks#1{\protected@xdef\@thanks{\@thanks
        \protect\footnotetext{#1}}}
\makeatother
\author{\bf \hypersetup{linkcolor=black} Harsh Nishant Lalai \bitslogo \hspace{0.4in}
        Aashish Anantha Ramakrishnan \psulogo \\
        \bf \hypersetup{linkcolor=black} Raj Sanjay Shah \gtlogo \hspace{0.4in}
        Dongwon Lee \psulogo\\
        Birla Institute of Technology and Science, Pilani \bitslogo\\
        The Pennsylvania State University \psulogo \hspace{0.4in} Georgia Institute of Technology \gtlogo \\
        \thanks{Email: f20212665@goa.bits-pilani.ac.in, \{aza6352, dongwon\}@psu.edu, rajsanjayshah@gatech.edu}
        }

\begin{document}
\maketitle
\begin{abstract}



With the rapid growth of Large Language Models (LLMs), safeguarding textual content against unauthorized use is crucial. Watermarking offers a vital solution, protecting both - LLM-generated and plain text sources. This paper presents a unified overview of different perspectives behind designing watermarking techniques through a comprehensive survey of the research literature. Our work has two key advantages: (1) We analyze research based on the specific intentions behind different watermarking techniques, evaluation datasets used, and watermarking addition and removal methods to construct a cohesive taxonomy. (2) We highlight the gaps and open challenges in text watermarking to promote research protecting text authorship. This extensive coverage and detailed analysis sets our work apart, outlining the evolving landscape of text watermarking in Language Models.

\end{abstract}

\section{Introduction}
Large Language Models (LLMs) 
can mimic human-like comprehension and text generation \cite{zheng2024judging}. Consequently, judging whether a text is authored by a human or generated by an LLM is challenging. This issue is highlighted by the recent lawsuit of The New York Times against OpenAI and Microsoft concerning the use of their articles as training data for AI models, emphasizing the need for effective methods to identify and safeguard digital content ownership \cite{initial_suit}.

\textbf{Text Watermarking} provides key solutions to protect intellectual property rights, identify ownership, and keep track of digital content. These techniques embed imperceptible signals or 
identifiers within digital text documents, which are then used to track the document's origins \cite{jalil2009review,kamaruddin2018review}. In particular, they aid in tracking the different production sources of text, both human-written and LLM-generated, helping prevent their unauthorized use without the owner's consent. 



Given this increasing research focus on watermarking techniques, it is important to review various methods, their applications, strengths and limitations. This includes systematically categorizing current research literature and highlighting key open challenges. The following contributions of our work distinguish it from previous surveys:


\begin{itemize}[leftmargin=*]
    \setlength\itemsep{0em}
    \setlength\parskip{0em}
    \setlength\parsep{0em}
    \item \textbf{Taxonomy Construction:} We seek to help future researchers navigate text-watermarking by categorizing various techniques and methods. Unlike traditional surveys, our paper aims to use the constructed taxonomy to provide an up-to-date list of research challenges for the text watermarking field instead of the most up-to-date survey of the field. For this task, we focus on \emph{application-driven intentions, evaluation data sources, and watermark addition methods}. We also enlist potential adversarial attacks against these methods to caution readers. 

    \item \textbf{Open Challenge Identification:} Next, we describe open challenges and gaps in current research efforts. These span rigorous testing of methods against diverse de-watermarking attacks, the establishment of standardized benchmarks for appropriate method efficacy comparison, expansion of the downstream NLP tasks used for evaluation, understanding how watermarking impacts language model factuality and utility, the interpretability of watermarking techniques by detailed descriptions and visual aids, developing personalized watermarking techniques, and lastly, collective watermark creation in multi-agent settings.
\end{itemize}
Our work aims to enable researchers to recognize emerging trends and areas for improvement in text watermarking research. We facilitate this goal by creating a systematic and comprehensive taxonomy of text watermarking.

\section{Taxonomy of Text Watermarking}

To help researchers navigate the field of text watermarking, we cluster various techniques and methods based on key commonalities. For this categorization, we focus on \emph{intentions that are application-driven, data sources for model evaluation, watermark addition methods, and method-specific adversarial attacks}. In our taxonomy creation, we allow techniques to belong to multiple categories and show how different techniques relate across multiple dimensions, making it easier to navigate the field.

\subsection{Intention}

Methods for embedding textual identifiers to watermark differ based on a user's desired features, the user's role (developer vs end-user, etc.), and primary application-driven needs. We categorize watermarking techniques based on the \emph{end user's intention} into three types: \textit{Text Quality, Output Distribution}, and \textit{Model Ownership Verification}.

\subsubsection{Text Quality} \label{sec:Text Quality}

Maintaining the quality and utility of the generated text post-watermarking is a desired goal of any watermarking methodology. However, research works differ on definitions of quality and mainly proxy output quality with (1) \emph{generation perplexity (uncertainty)} and (2) \emph{semantic relatedness of watermarked and un-watermarked generations}.


\paragraph{Minimizing impact on Perplexity} A model's confidence in its generations, measured through the weighted sum of individual token log probabilities in a sequence is known as Perplexity. A lower perplexity indicates that the model is more confident and accurate in its predictions, while a higher perplexity suggests more significant uncertainty and less accurate predictions. Perplexity is the only intrinsic measure of model uncertainty \cite{magnusson2023paloma}, and thus, a popular measure of quality among researchers.

\begin{figure}[t]
\centering
\includegraphics[trim={1cm 0cm 1cm 1cm}, width=0.41\textwidth]{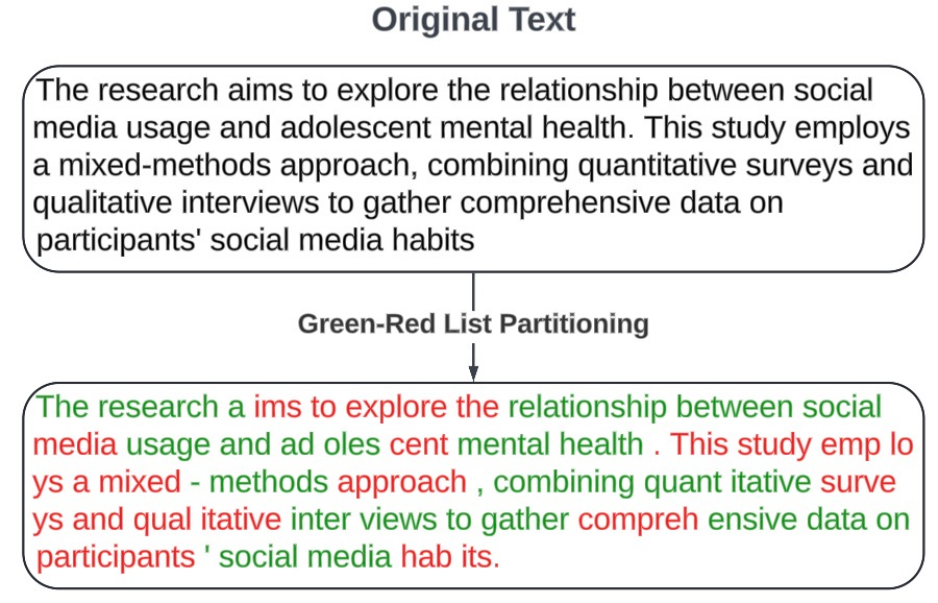}
\caption{An example of green-red list grouping of texts \cite{kirchenbauer2023watermark}.}
\label{fig:greenred}
\end{figure}

Watermarking techniques like using green-red list rules (refer to figure \ref{fig:greenred}) trade-off the ability to detect LLM-generated text and the utility of the output text. For a given text, the greater the proportion of green tokens from the total tokens, the lesser the chance of the text being written by humans. A parallel aim is to reduce all of the other "generic" text's perplexity while enforcing a more frequent generation of white-listed "green" words. 
Controlling entropy levels ensures that watermarked text maintains a quality similar to non-watermarked text. Studies have effectively operationalized this technique in diverse ways for authorship detection while maintaining high text quality \cite{kirchenbauer2023watermark,zhao2023provable,takezawa2023necessary}. For example, soft watermarking promotes green list use for high-entropy (rare) tokens while minimally affecting low-entropy (common) tokens \cite{kirchenbauer2023watermark,lee2023wrote, ren-etal-2024-subtle}, ensuring that watermark is undetectable (soft) to an observer. In another example,\citet{takezawa2023necessary} recommend a lower watermark strength for longer texts for quality. Some techniques only alter text appearance, for example, change "e" to "\'e", rather than modifying the content to have no perplexity impact \cite{464718,POR20121075,sato2023embarrassingly}. 
\newcommand{\cmark}{\textcolor{green!80!black}{\ding{51}}}
\newcommand{\xmark}{\textcolor{red}{\ding{55}}}

\begin{table}[!h]
    \centering
    \caption{Overview of watermarking techniques using semantic relatedness. \textit{Struct}: Maintains Structure, \textit{Word repl}: Synonym/ Spelling based word replacement techniques, \textit{Dep. trees}: Dependency trees, \textit{Syn. trees}: Syntax trees, \textit{POS}: Part-of-speech tagging, \textit{Lat-rep}: Latent representation based methods.}
    \label{tab:sem_related}
    \resizebox{\columnwidth}{!}{%
    
    \begin{tabular}{lcccccc}
    \hline
        Work &  Struct  & Word   & Dep.   & Syn.  & POS  & Lat. \\ 
         &   & repl.  &  trees  &  trees &   &  rep.  \\
        
        \hline
        \cite{10.1145/1161366.1161397} & \cmark & \cmark & \xmark & \xmark & \xmark & \cmark  \\
        \cite{meral2009natural} & \xmark & \xmark & \xmark & \cmark & \xmark & \xmark  \\
        \cite{abdelnabi2021adversarial} & \cmark & \xmark & \xmark & \xmark & \xmark & \cmark  \\

        \cite{yang2022tracing} & \cmark & \cmark & \xmark & \xmark & \xmark & \cmark  \\
        \cite{he2022protecting} & \cmark & \cmark & \cmark & \xmark & \cmark & \cmark  \\
        \cite{he2022cater} & \cmark & \cmark & \cmark & \cmark & \cmark & \xmark \\
        
        \cite{yoo2023robust} & \xmark & \xmark & \cmark & \xmark & \xmark & \xmark  \\
        \cite{yang2023watermarking} & \cmark & \cmark & \xmark & \xmark & \xmark & \cmark  \\
        \cite{munyer2023deeptextmark} & \cmark & \cmark & \xmark & \xmark & \xmark & \cmark  \\
        
        \cite{fu2024watermarking} & \xmark & \xmark & \xmark & \xmark & \xmark & \cmark  \\

        \cite{hoang2024less} & \xmark & \xmark & \xmark & \xmark & \cmark & \xmark  \\
        \hline
    \end{tabular}
    }
    
\end{table}

\paragraph{Semantic Relatedness} Refers to how closely words, phrases, or sentences of the watermarked output are similar to the original clean output. One way of maintaining input semantics is by embedding both input and output sentences into a semantic space and minimizing the distance between them \cite{abdelnabi2021adversarial, zhang2023remark}. \citet{yang2022tracing} use the BERT model to suggest substitution candidates, while other works use synonyms and spelling replacements to have minimum impact on semantic relatedness. \citet{fu2024watermarking} use the input context to extract semantically related tokens, measured by word vector similarity to the source.
In more nuanced domains like code generation, the preservation of semantics has been achieved by changing variable names \cite{li2023towards,yang2023towards}. Table \ref{tab:sem_related} provides an overview of the watermarking techniques using semantic relatedness.
\citet{chen2023x}  split synonyms or semantically similar words between "green" and "red" lists. This excludes the possibility of all suitable alternatives being placed in the same list, ensuring that if one synonym is discouraged, another remains available. This allows LLMs to maintain their original articulation ability.
Alternatively, \citet{liu2024adaptive} use a semantic-based logits scaling vector extraction approach. This method adjusts the logits based on the meaning of the previously generated text, ensuring that the watermark perturbations align with the original text's meaning. \citet{li2024resilient} involve reordering operations and code formatting changes, such that they do not alter the functionality or degrade the quality of the code. 


\subsubsection{Similar Output Distribution}
Ensuring that the word distribution in watermarked text or LLM-generated output closely resembles that of the original text is essential for providing a natural experience to the end user. This is often operationalized as re-weighting strategies that adjust the probabilities of select words during text generation such that the overall distribution of words remains consistent with the original \cite{hu2023unbiased,wu2023dipmark}. 

\citet{hu2023unbiased} and \citet{wu2023dipmark} focus on creating stealthy watermarks that remain imperceptible and avoid introducing noticeable biases. By adjusting the output logits of LLMs, these methods preserve the original text distribution and minimize the likelihood of detection. In some methods, this is done by systematically rearranging the words (permutation) in the vocabulary set to find optimal combinations that maintain the inherent symmetry of the original distribution \cite{wu2023dipmark}.
This method exploits the mathematical property of symmetry in permutations, where different arrangements can still produce the same statistical distribution, allowing for flexibility in embedding watermarks without altering the natural flow of the text.

\subsubsection{Model Ownership Verification}

Model ownership verification techniques use watermarks to safeguard against adversaries by helping model creators prove ownership, even if adversaries attempt to emulate the model’s functionality. For an adversary, emulating LLM behavior requires understanding the workings of a model. An adversary's goals include model extraction - where they seek to exploit or verify the properties of an LLM and recreate the model by extensively querying it.
Attackers can have varying levels of access to the model: \textit{black-box access} (input queries and receive outputs without internal knowledge), \textit{white-box access} (full knowledge of architecture, parameters, and training data), and \textit{gray-box access} (partial knowledge, such as architecture without parameters).

\begin{table}[!h]
    \centering
    \caption{Overview of watermarking techniques for Model Ownership Verification. \textit{Trigger Sets}: Watermark Location Indicators, \textit{Msg Inj}: Message Injection, \textit{App}: Change in appearance.}
    \label{tab:attack}
    \resizebox{\columnwidth}{!}{%
    
    \begin{tabular}{lcccc}
    \hline
        Work & Trigger & Secret & Msg & App. \\ 
         & Sets & Keys & Inj    \\
        
        \hline

        \cite{464718} & \xmark & \xmark & \xmark & \cmark  \\
        \cite{10.1007/3-540-45496-9_14} & \xmark & \cmark & \xmark  & \xmark \\
        \cite{POR20121075} & \xmark & \xmark & \xmark & \cmark  \\

        \cite{dai2022deephider} & \cmark & \cmark & \xmark  & \xmark \\

        \cite{peng2023you} & \cmark & \xmark & \xmark & \xmark  \\

        \cite{tang2023did} & \cmark & \xmark & \xmark & \xmark \\

        \cite{zhang2023remark} & \xmark & \xmark & \cmark & \xmark  \\
        \cite{cryptoeprint:2023/1661} & \xmark & \cmark & \cmark & \xmark  \\
        \cite{kuditipudi2023robust} & \xmark & \xmark & \cmark & \xmark  \\
        \cite{sato2023embarrassingly} & \xmark & \xmark & \xmark & \cmark  \\
        \cite{zhao2023provable} & \xmark & \cmark & \xmark & \xmark  \\
        \cite{zhao2023protecting} & \xmark & \cmark & \cmark & \xmark  \\
        \cite{liu2023watermarking} & \cmark & \xmark & \xmark & \xmark  \\
        \cite{shao2024explanation} & \cmark & \xmark & \xmark & \xmark \\
        \cite{qu2024provably} & \xmark & \cmark & \cmark & \xmark  \\

        \hline
    \end{tabular}
    }
    
\end{table}


The attack conditions define the environment and constraints under which the attack is conducted. These conditions include resource constraints (computational resources like processing power, memory, and time), access constraints (black box, white box, or gray box), knowledge assumptions (information the attacker has about the model, including architecture, training data, or defense mechanisms), detection and evasion (avoiding detection if the model has monitoring systems), and performance metrics (criteria for evaluating attack success, such as accuracy of model extraction, watermark detection consistency, or successful adversarial perturbations).

Combating attackers often requires a technique with minimal false positives, i.e., the unauthorized emulation of LLMs is easily detected. For model ownership verification, techniques like trigger sets rely on the predictability of specific outputs given exact inputs. Trigger sets are specific inputs designed to activate watermarks embedded within a model or dataset \cite{dai2022deephider,peng2023you,liu2023watermarking,tang2023did}. \citet{dai2022deephider} uses secret keys for embedding and detecting watermarks, while others use lexical features for watermarking. 



Injecting secret signals/messages/signatures in the watermark generation process is also used for verification \cite{zhao2023protecting,zhang2023remark,cryptoeprint:2023/1661,qu2024provably,kuditipudi2023robust, wang2023towards, zhou2024bileve}. \citet{wang2023towards}, \citet{qu2024provably} and \citet{guan2024codeip} embed a multi-bit watermark into the output logits of LLMs which can indicate the model's identity or version, user information about who prompted the generation, and timestamp or contextual details relevant for tracking or verification. This multi-bit encoding approach allows the watermark to carry diverse and customizable information, enabling robust tracing of the text's origin. \citet{zhao2023provable} use a secret key to vary the green list's length, allowing personalized watermarking. 



\subsection{Watermark Addition}

\begin{figure}[htp]
\centering
\includegraphics[trim={1cm 0cm 2cm 1cm}, width=0.41\textwidth]{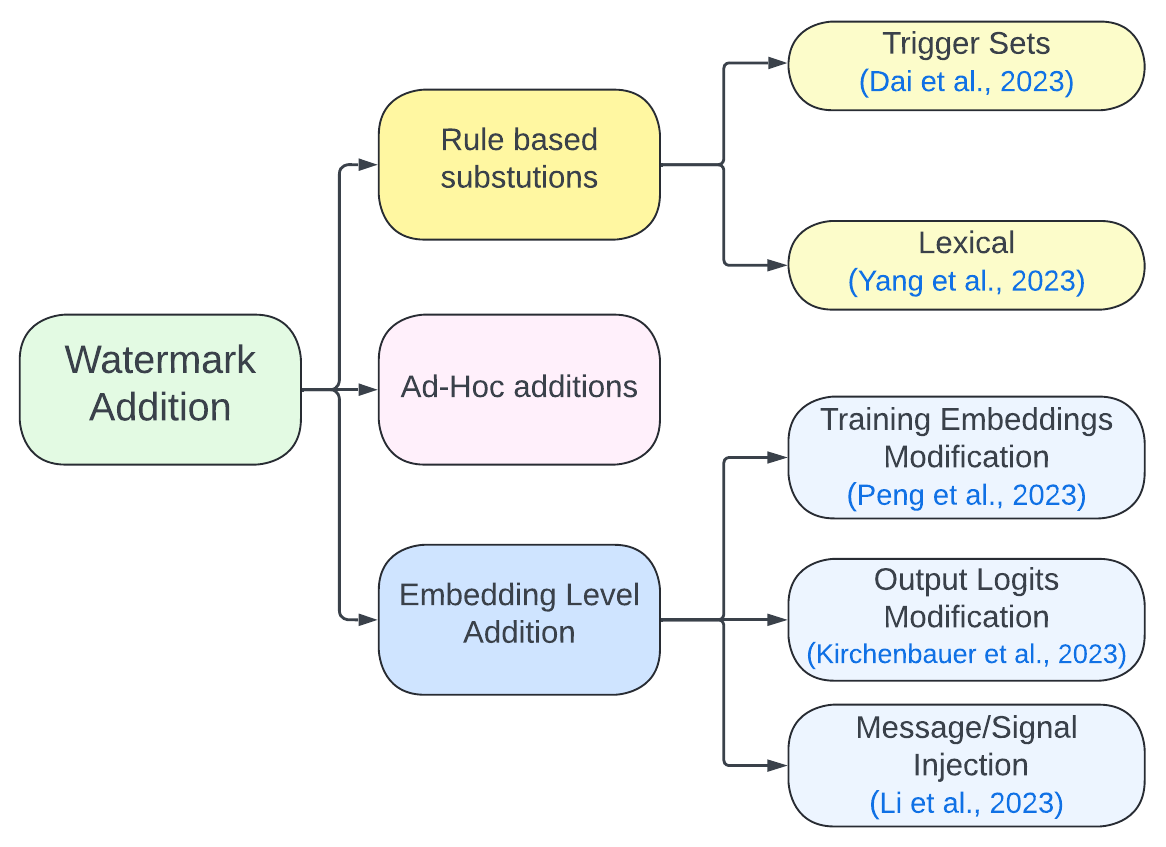}
\caption{Sub-categorization of various Watermark Additions.}
\label{fig:introductionadd}
\end{figure}

Often, the same watermarking methods work for different user intentions. Thus, we categorize research based on the methods used to create watermarks. As shown in Figure \ref{fig:introductionadd}, techniques primarily fall into three distinct categories: \emph{Rule-Based Substitutions, Embedding-Level Addition, and Ad-Hoc Addition}.

\subsubsection{Rule Based Substitution}

In rule-based substitution techniques, certain elements are replaced in the text based on specific rules or patterns while preserving the overall structure and semantics of the text. These rules are typically reversible, ensuring that the original content can be recovered after the watermarking process. Rule Based Substitution techniques can be further divided into two categories, namely \emph{Trigger set-based and Lexical methods}.

\paragraph{Trigger Sets} Refers to specific conditions or patterns that activate or reveal the watermark embedded within the text. Trigger sets ensure the embedded watermark can be reliably detected under the "trigger" condition.
Trigger sets have been operationalized in many ways; for example, \citet{dai2022deephider} create trigger sets for multi-task learning (for example, a three-way classification problem) by selecting a small number of samples from different classes to obtain LLM prediction probabilities over all categories. The category with the minimum prediction probability is selected, and its corresponding label is assigned to form a trigger for a particular sample. Similarly, \citet{liu2023watermarking} create trigger sets at different text granularity, namely character, word, and sentence levels, by adding or appending a character/sentence/word within text data for multi-task learning. Other types of trigger sets include word-level \cite{peng2023you} and style-level\cite{tang2023did} triggers. Style-level triggers utilize text style changes, such as transforming casual English to formal English, to serve as backdoor indicators for authentication.

\begin{figure}[t]
\centering
\includegraphics[trim={1cm 0cm 2cm 1cm}, width=0.41\textwidth]{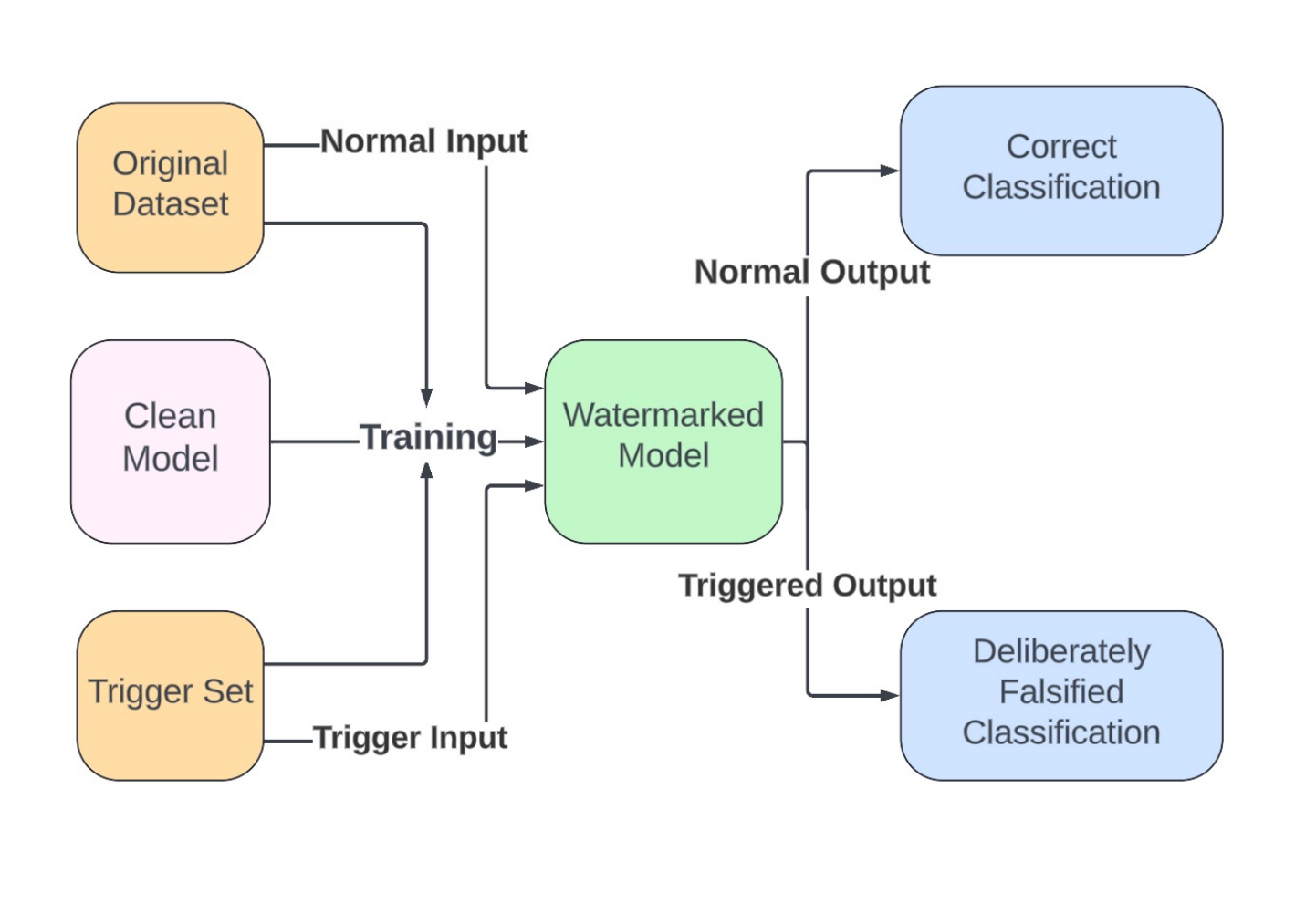}
\caption{Operationalization of some Trigger-set based watermarks. Here, the original model is trained with the trigger set, which modifies the input to change the class of the output deliberately \cite{liu2023watermarking} or change the output for the same input \cite{dai2022deephider}.}
\label{fig:triggerset}
\end{figure}
\paragraph{Lexical substitution} These techniques deterministically replace words and phrases with alternative lexical units while maintaining content coherence and semantics. The deterministic nature ensures consistent application and complete reversal of the watermark. 
A straightforward operationalization of lexical replacement is based on semantic preservation, which includes synonym replacement using wordnet \cite{he2022protecting, yang2023watermarking}, spelling variant replacement between US and UK spellings \cite{10.1145/1161366.1161397}, model-in-the-loop semantic similarity based search between candidate replacements and original sentence\cite{munyer2023deeptextmark, yang2022tracing}. In a nuanced domain like code generation, methods have looked at reordering operations and code formatting changes, which do not alter the functionality \cite{li2024resilient}



\subsubsection{Embedding-level Addition}
Watermarking techniques can be distinguished based on \emph{how} the watermarks are embedded. These broadly include \textit{Train-time watermarking, Output Logits Modification, and Message/Signal Injection}.

\paragraph{Train-time watermarking} As the name suggests, this method embeds the watermark during training time. \citet{peng2023you} select a group of moderate-frequency words from a general text corpus to form a trigger set, then select a target word as the watermark and insert it into the latent representations of texts containing trigger words as the backdoor. 

\paragraph{Output Logits Modification} The output logits of LLMs are unnormalized scores assigned to each token before applying the softmax function to generate probabilities. These probabilities reflect the model's confidence in predicting tokens. Logits determine the model's token prediction (where the highest logit determines the predicted token), training (comparing logits with actual labels to compute the loss), and interpreting model behavior by highlighting token importance. These methods modify the post-softmax distributions over the model's vocabulary.

A popular example of an Output Logit Modification watermarking is the use of green-red lists \cite{kirchenbauer2023watermark, lee2023wrote, zhao2023provable, takezawa2023necessary, fu2024watermarking, ren2023robust, wu2023dipmark, chen2023x}, methods typically vary in the choice of high/low entropy tokens to add to the green list, size in the watermark (number of bits), injection of complex vs soft watermark, discarding low probability tokens, ensuring semantically similar words are distributed across green and red lists.

Apart from the techniques above, other methods involve injecting secret signals into the probability vector of the decoding steps for each target token \cite{zhao2023protecting}. \citet{liu2023semantic, liu2024adaptive} dynamically determine the logits to watermark with the help of semantics of all preceding tokens. Specifically, \citet{liu2023semantic} utilizes another embedding LLM to generate semantic embeddings for all preceding tokens, and then these semantic embeddings are transformed into the watermark logits through their trained watermark model. Building from the idea of secret signals, \citet{cryptoeprint:2023/1661} use cryptographic digital signatures through a private key in text generation, which is then detected using a public key. Similarly, research also explores embedding multi-bit information into output logits \cite{qu2024provably, wang2023towards, guan2024codeip}.


\paragraph{Message/Signal Injection}
Watermarks can be encoded in the text itself or used by functions to map values with the text to be watermarked. 
These procedures involve the injection of messages, signals, or bit strings in the latent space of the text created by the encoders\cite{wang2023towards, guan2024codeip}. For example, \citet{li2023towards} tasks the representations of the abstract syntax tree (AST) tokens as input to predict modified variable names with encoded bit strings and \citet{yang2023towards,li2023towards} encode identifier bit strings into the source code, without affecting the usage and semantics of the code. They perform transformations on an AST-based intermediate representation that enables unified transformations across different programming languages involving the changes in the expression, statement, and block attributes. \citet{zhang2023remark} use linear combinations within this latent space to add a simple message to the embedded text. The decoder then converts it back into plain text with minor modifications resulting from the added message. A similar process is implemented to encode bit strings containing information like user ID and generation date \cite{qu2024provably}. \citet{zhou2024bileve} injects coarse-grained and fine-grained signals (signatures) into the text during generation. The coarse-grained level utilizes statistical signals to detect watermark presence, while the fine-grained level embeds content-dependent signature bits for verifying content integrity.

\subsubsection{Ad-Hoc Addition }


Unlike popular watermarking methods like rule-based substitutions, which have strict, global definitions for modifications within a sentence (like synonym/spelling replacement and triggers), the ad-hoc addition methods use task-specific local guidelines for changes to the sentence structure. We bucket these methods into \emph{Ad-Hoc addition methods} and list a few relevant methods. 

First, \citet{POR20121075, sato2023embarrassingly} embed watermarks by inserting Unicode spaces in the text. \citet{sato2023embarrassingly} introduce three methods: \emph{WhiteMark} replaces whitespace with alternate Unicode spaces (e.g., U+0020 to U+2004), \emph{VariantMark} uses Unicode variation selectors to embed messages in Chinese, Japanese, and Korean texts by substituting characters with variants, and \emph{PrintMark} alters text appearance for printed media through ligatures, varied spaces, and character variants. Another work introduces three unique syntax transformations for message encoding— Adjunct Movement, Clefting, and Passivization \cite{10.1007/3-540-45496-9_14}. For instance, Adjunct Movement involves relocating adjuncts within a sentence, as demonstrated by the variability in positioning the word 'quickly' in "She quickly finished her homework." Clefting highlights a specific clause, typically the subject, such as transforming "The chef cooked a delicious meal" into "It was the chef who cooked a delicious meal" to emphasize 'the chef.' Passivization, however, changes active sentences with transitive verbs into passive voice, transforming "The teacher graded the exams" into "The exams were graded by the teacher." 
\citet{sun2023codemark} apply semantic-preserving code transformations by modifying operators.

\paragraph{Overlapping categories}
Some papers span multiple categories within the taxonomy. For example, \citet{zhao2023provable} and \citet{sato2023embarrassingly} address both Text Quality and Model Ownership Verification, preserving readability while ensuring ownership with detectable markers. Similarly, the work by \citet{yang2023towards} is listed under both message injection and embedding-level addition, as it involves injecting watermarks as messages within the embedding space. Similarly, research presented by \citet{peng2023you} falls under both embedding-level additions and trigger-set because it uses modified embeddings activated by specific triggers to enhance watermark detectability.
These dual listings reflect the flexibility provided by these techniques, addressing overlapping goals. In contrast to prior surveys with limited focus areas, we believe maintaining separate but overlapping categories helps clarify distinct objectives and evaluation criteria across multiple dimensions, ensuring comprehensive coverage of the taxonomy.


\begin{table*}[t]
\caption{Datasets used in the evaluation of watermarking techniques. \textbf{Bold} indicates the most used dataset(s) for a particular downstream NLP task and the respective works using the dataset.}
\label{dataset_table1}
\centering
\renewcommand{\arraystretch}{1.3}
\resizebox{1\textwidth}{!}{%

\begin{tabular}{>{\raggedright\arraybackslash}p{3cm} >{\raggedright\arraybackslash}p{11.5cm} >{\raggedright\arraybackslash}p{7cm}}
\toprule
\textbf{Downstream Task} & \textbf{Dataset Name} & \textbf{Papers} \\ \midrule

Text Completion & \textbf{Colossal Clean Crawled Corpus (C4) \cite{raffel2020exploring}}, Dbpedia Class \cite{10.1007/978-3-540-76298-0_52}, WikiText-2 \cite{merity2016pointer} &  \textbf{\citet{kirchenbauer2023watermark}, \citet{kuditipudi2023robust}, \citet{liu2023private}, \citet{munyer2023deeptextmark}, 
\citet{yoo2023advancing}, \citet{liu2023semantic}, \citet{cryptoeprint:2023/1661}, \citet{ren2023robust}, \citet{hou2023semstamp}, \citet{qu2024provably}, \citet{wang2023towards}, \citet{liu2024adaptive}, \citet{chen2023x}, \citet{mao2024watermark}, \citet{chang2024postmark}}, \citet{zhou2024bileve}\\ \midrule

Post-watermark text similarity analysis & \textbf{WikiText-2, Workshop on Statistical Machine Translation (WMT14) \cite{bojar-etal-2014-findings}}, Internet Movie Database (IMDb) \cite{maas-etal-2011-learning}, AgNews \cite{zhang2015character}, Dracula, Pride and Prejudice, Wuthering Heights \cite{gerlach2020standardized}, CNN/Daily Mail \cite{nallapati2016abstractive},  Human ChatGPT Comparison Corpus (HC3) \cite{guo2023close}, C4,  Reuters Corpus \cite{lewis2004rcv1}, ChatGPT Abstract \cite{sivesind_2023}, Human Abstract \cite{sivesind_2023} & \textbf{\citet{yang2022tracing}, \citet{he2022protecting}, \citet{he2022cater}, \citet{yoo2023robust}}, \textbf{\citet{sato2023embarrassingly}},
\textbf{\citet{zhang2023remark}},
\citet{yang2023watermarking}, \citet{Topkara2006WordsAN} \\ \midrule

Machine Translation & \textbf{WMT14}, IWSTL14 \cite{CettoloNiehuesStueker2014_1000166288} & \textbf{\citet{zhao2023protecting}, \citet{wu2023dipmark}, \citet{hu2023unbiased}, \citet{takezawa2023necessary}} \\ \midrule

Text Summarisation & \textbf{CNN/Daily Mail}, Extreme Summarization (XSUM) \cite{narayan2018don}, Data Record to Text Generation (DART) \cite{nan-etal-2021-dart} , WebNLG \cite{gardent-etal-2017-creating} & \textbf{\citet{fu2024watermarking}, \citet{wu2023dipmark}, \citet{hu2023unbiased}} \\ \midrule

Code Generation & \textbf{CodeSearchNet (CSN) \cite{husain2019codesearchnet}}, HUMANEVAL \cite{chen2021evaluating}, Mostly Basic Python Programming (MBPP), MBXP \cite{athiwaratkun2023multilingual}, DS-1000 \cite{lai2023ds}, APPS \cite{hendrycks2021measuring} & \textbf{\citet{li2023towards}, \citet{yang2023towards}, \citet{guan2024codeip}},
\citet{lee2023wrote}, \citet{li2024resilient}, \citet{mao2024watermark}\\ \midrule

Question Answering & \textbf{OpenGen \cite{krishna2024paraphrasing}, Long Form Question Answering (LFQA) \cite{krishna2024paraphrasing}}, TruthfulQA \cite{lin2021truthfulqa} & \textbf{\citet{zhao2023provable}, \citet{yoo2023advancing}, \citet{qu2024provably}, \citet{zhou2024bileve}, \citet{chang2024postmark}},
\citet{chen2023x}\\ \midrule

Story Generation & \textbf{ROCstories \cite{mostafazadeh2016corpus}} & \textbf{\citet{zhao2023protecting}} \\ \midrule

Text Classification & \textbf{Stanford Sentiment Treebank (SST) \cite{socher-etal-2013-recursive}, AgNews, Microsoft News Dataset (MIND) \cite{wu-etal-2020-mind}, Enron Spam \cite{inproceedings}} & \textbf{\citet{peng2023you}} \\ \bottomrule

\end{tabular}
}
\end{table*}


\subsection{Evaluation} \label{sec:Evaluation}

A wide variety of datasets have been used to evaluate the performance of watermarking approaches, limiting our ability to extract generalized conclusions about their performance. Different benchmarks focus on selected downstream tasks to validate watermarking capabilities, and we provide a detailed breakdown of the datasets utilized in Table \ref{dataset_table1}. We observe many evaluation datasets focusing on text completion and post-watermarking text similarity tasks. The downstream task descriptions are provided below.

\subsubsection*{Downstream Task descriptions}
\paragraph{Text Completion Task} This task involves giving the LLM a portion of text from the dataset as a prompt and then asking it to complete the text. The generated completion is then compared with the human completion or the portion of the dataset not provided as the prompt.

\paragraph{Post-watermark text similarity analysis} In this task, given an initial text \(X\), watermarking is applied to \(X\) to produce a modified text \(X'\). An example could be a rule-based substitution with synonyms or spelling replacements. The comparison is then made between \(X\) and \(X'\) based on distinctions in length, semantics, and other linguistic features.

\paragraph{Other Downstream Tasks} For these tasks, given the same initial prompt \(X\), the LLM's generated response \(Y\) (before watermarking) is compared with the response \(Y'\) (after watermarking).

\subsection{Adversarial attacks on watermarking techniques}

Malicious and adversarial actors seek to misuse LLM technology and bypass watermarks to avoid being distinguished from rightful owners. To promote research into protecting intellectual property rights, we extend suggestions from \citet{kirchenbauer2023watermark} to describe de-watermarking methods, i.e., adversarial attacks on text watermarking, into three categories:



\begin{figure}[t]
\centering
\includegraphics[trim={0cm 0cm 0cm 0cm}, width=0.5\textwidth]{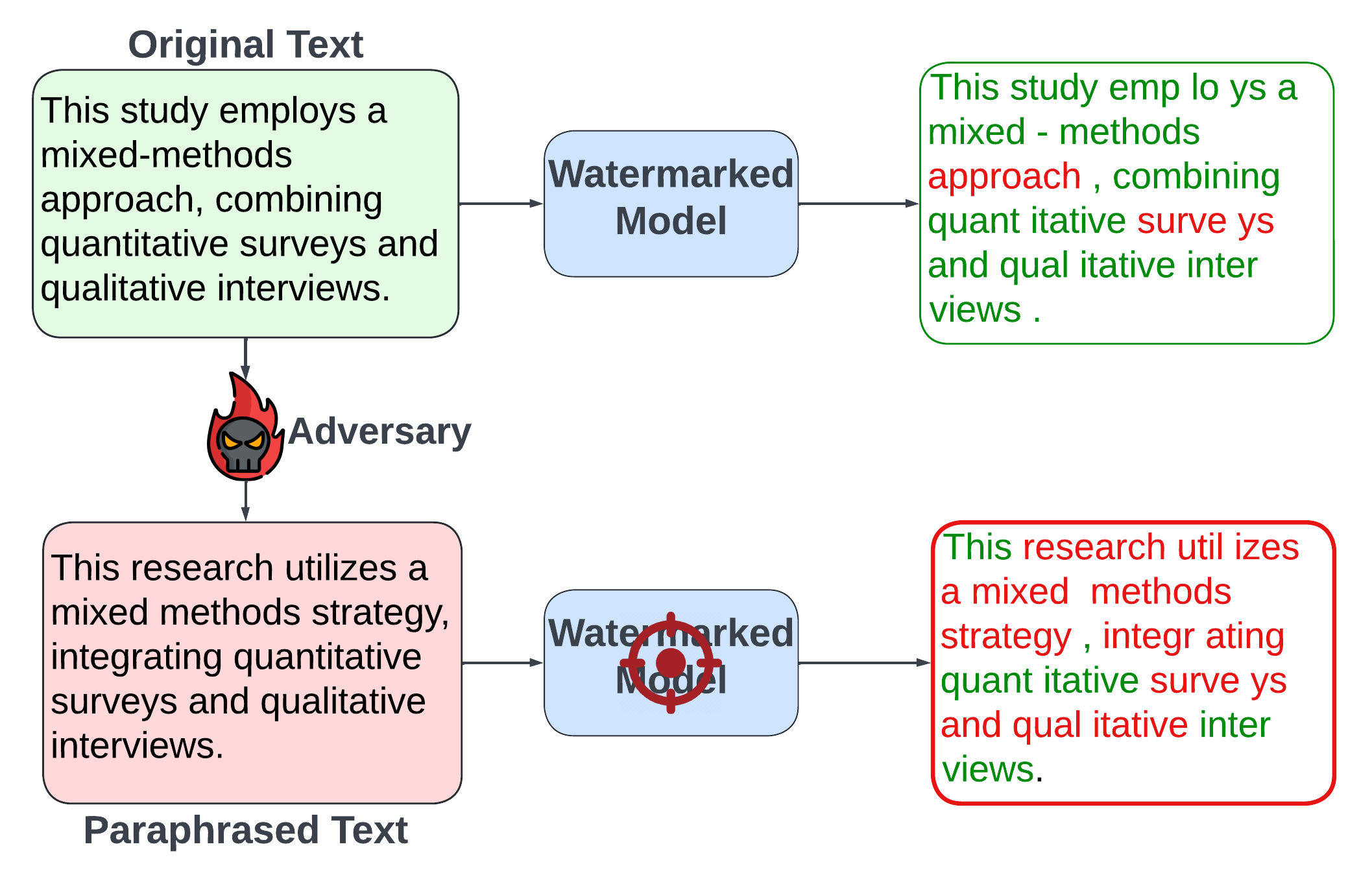}
\caption{An example of an adversary performing a de-watermarking attack on a green-red list-based watermarking technique. The original partitioning contains a higher proportion of green tokens than the partitioning after adversarial paraphrasing.}
\label{fig:adv}
\end{figure}

\begin{enumerate}[leftmargin=*]
    \item \textbf{Text insertion attacks} involve adding additional tokens or text segments to the original output of a watermarked LLM generation. For example, on watermarking methods with green-red lists \cite{kirchenbauer2023watermark, zhao2023protecting, takezawa2023necessary}, an attacker could add additional tokens from the red list, leading to the obfuscation of the watermarking method. Another variant of text insertion attacks includes copy-paste attacks \cite{qu2024provably}, where adversaries insert copied text from external sources into the watermarked output. This approach can dilute the effect of the watermark by embedding non-watermarked content within the watermarked text, further complicating watermark detection and attribution.
    
    \item \textbf{Text deletion attacks} involve the removal of tokens or text segments from the original watermarked output of an LLM and modifying the rest of the tokens to fit the output. Returning to the example of green-red list methodologies, this means removing some of the green list tokens from the output and modifying the red list tokens in the output \cite{kirchenbauer2023watermark}. These techniques often require knowledge of the vocabularies belonging to each of the two lists in green-red lists.
    
    \item \textbf{Text substitution attacks} entail replacing specific tokens or text segments in the watermarked output while preserving its overall meaning. Attackers perform tokenization attacks by paraphrasing text \cite{ren2023robust}, misspelling words, or replacing characters like newline (\textbackslash n); increasing red list tokens, and evading green-red list watermarking \cite{kirchenbauer2023watermark}. These also include Homoglyph attacks: attacks that exploit Unicode characters that look similar but have different IDs, leading to variation from expected tokenization (e.g., "Lighthouse" becomes nine tokens with Cyrillic characters). Generative attacks leverage LLMs' context learning to manipulate the output predictably, such as adding emojis after each token or replacing characters to disrupt watermark detection \cite{kirchenbauer2023watermark}.
\end{enumerate}


\section{Discussion and Open Challenge} \label{discussion}


Our taxonomy-driven categorization of the research space exposes the open challenges to watermarking and outlines "good to have" criteria while developing new techniques to protect intellectual property ownership. They are as follows:
    \paragraph{Resilience to adversarial attacks} One of the critical challenges in the field is the lack of \emph{comprehensive} evaluation of techniques against a diverse range of de-watermarking attacks. While many researchers focus on developing robust techniques, there is often insufficient emphasis on systematic red-teaming of these methods against multiple attacking scenarios.
    \paragraph{Standardization of evaluation benchmarks} There is a need for standardized benchmarks and evaluation metrics to ensure fair and consistent comparison between different watermarking techniques. Table \ref{dataset_table1} shows how evaluation datasets differ in the literature for the same downstream task, reflecting this necessity.

    \paragraph{Impact on LLM output factuality} Watermarks modify the model output distributions; techniques that are robust to de-watermarking often have greater variations in watermarked outputs compared to clean outputs, leading to a potential trade-off between de-watermarking and LLM factuality. Despite this potential trade-off, there is a lack of analysis on how watermarking techniques affect the output inaccuracies or hallucinations. After training or fine-tuning LLMs with specific watermarking techniques, there is often insufficient examination of whether these methods introduce or exacerbate inaccuracies. We advocate for factuality evaluations post-watermarking. 

        
    \paragraph{Enhanced Interpretability} Drawing upon security and privacy literature \cite{kumar2024ethics}, we ask the community to establish privacy norms for LLM watermarking. We envision this to be similar to model cards, which describe the degree of security provided by particular methods against malicious actors.

    \paragraph{Human-centered watermarking} We urge the community to work on the human perception of LLMs when interacting with different safety principles. User perception of LLMs may change with differences in output distributions. Furthermore, safety practices may enable AI acceptance and adoption among the masses.

    \paragraph{Personalized watermarking} Developing personalized watermarking approaches that cater to individual user preferences, contexts, or specific organizational policies is important. This requires adaptive techniques that can embed user- or context-specific identifiers while respecting privacy, personalization preferences, and practical constraints.

    \paragraph{Multi-agent Watermarking} Current watermarking methods do not effectively support scenarios where multiple LLM agents sequentially or collaboratively contribute to generating text. Specifically, existing approaches fail to maintain watermark integrity and detectability when multiple distinct watermarks are stacked or embedded in succession. An unresolved challenge is developing watermarking techniques that enable each agent to reliably detect its own watermark and a collaboratively embedded watermark, even when multiple watermarks from different agents coexist within the same generated content.
    


\section{Conclusion} \label{conclusion}

In this paper, we analyze representative literature to provide a comprehensive taxonomy for digital watermarking techniques for both LLM-generated and human-written text. The taxonomy categorizes watermarking techniques using four primary categories, namely - intention of the method, data used for evaluation, watermark addition, and adversarial attacks. 

We identify and cluster existing watermarking methods, highlighting key open challenges and research gaps in the field. For every watermarking method, \emph{we advocate for establishing stronger evaluation paradigms: standardized datasets, resilience to adversarial attacks, impact on model utility and output actuality, enhanced interpretability, and human perception change upon the use of these such techniques.}
We envision this research as a reference for policymakers, safety practitioners, and end users, facilitating the adoption of robust digital watermarking practices and promoting responsible AI use.


\section{Limitations}

Limitations to our work are as follows: 
(1) We do not include detailed insights into metrics for success rate (accuracy of detecting watermarked texts), text quality (perplexity and semantics), NLP task-specific evaluation, and robustness (detectability of watermarks after removal attacks). However, we briefly describe the two types of metrics into watermarking success rates (intrinsic quality of watermarking) and model utility/ performance \ref{metrics for evaluation}. (2) Given the scope of this paper, we do not demonstrate the mathematical analysis of different watermarking techniques. We urge readers to refer to the original papers for the same (3) We do not cover all different task deployment scenarios for the watermarking techniques discussed. (4) Readers may perceive similarity between sections \ref{sec:Text Quality} (text quality) and \ref{sec:Evaluation} (evaluation), however, we wish to highlight the following distinction - in section \ref{sec:Evaluation} (evaluation), our focus is on the datasets that different papers use to evaluate their watermarking techniques whereas in section \ref{sec:Text Quality} (text quality), we look at how some papers have similar intentions or end goals of watermarking. While similar intentions often include many of the same downstream tasks as mentioned in sections  \ref{sec:Evaluation} (Text Completion Task, Post-watermark text similarity analysis, Other Downstream Tasks), and \ref{sec:Text Quality}, we advocate for the standardization of these tasks for definitive apples to apples comparison between techniques. (5) While we advocate for a standardized evaluation, we do not propose a framework for evaluating the effectiveness of watermarking techniques. (6) We focus primarily on watermarking and do not delve into the broader relationships between watermarking and adjunct fields, for example, steganography. However, additional information on these topics can be found in the appendix section \ref{stegano}.

\section{Ethical Considerations}
This paper reviews the challenges and opportunities of watermarking techniques in LLMs. Our work has many potential societal consequences, none of which must be specifically highlighted here. There are no major risks associated with conducting this review.

\section{Acknowledgments}
This work was partly supported by U.S. NSF awards \#1934782 and \#2114824.


\bibliography{acl_latex}

\begin{thebibliography}{81}
\providecommand{\natexlab}[1]{#1}

\bibitem[{Abdelnabi and Fritz(2021)}]{abdelnabi2021adversarial}
Sahar Abdelnabi and Mario Fritz. 2021.
\newblock Adversarial watermarking transformer: Towards tracing text provenance with data hiding.
\newblock In \emph{2021 IEEE Symposium on Security and Privacy (SP)}, pages 121--140. IEEE.

\bibitem[{Atallah et~al.(2001)Atallah, Raskin, Crogan, Hempelmann, Kerschbaum, Mohamed, and Naik}]{10.1007/3-540-45496-9_14}
Mikhail~J. Atallah, Victor Raskin, Michael Crogan, Christian Hempelmann, Florian Kerschbaum, Dina Mohamed, and Sanket Naik. 2001.
\newblock Natural language watermarking: Design, analysis, and a proof-of-concept implementation.
\newblock In \emph{Information Hiding}, pages 185--200, Berlin, Heidelberg. Springer Berlin Heidelberg.

\bibitem[{Athiwaratkun et~al.(2023)Athiwaratkun, Gouda, Wang, Li, Tian, Tan, Ahmad, Wang, Sun, Shang, Gonugondla, Ding, Kumar, Fulton, Farahani, Jain, Giaquinto, Qian, Ramanathan, Nallapati, Ray, Bhatia, Sengupta, Roth, and Xiang}]{athiwaratkun2023multilingual}
Ben Athiwaratkun, Sanjay~Krishna Gouda, Zijian Wang, Xiaopeng Li, Yuchen Tian, Ming Tan, Wasi~Uddin Ahmad, Shiqi Wang, Qing Sun, Mingyue Shang, Sujan~Kumar Gonugondla, Hantian Ding, Varun Kumar, Nathan Fulton, Arash Farahani, Siddhartha Jain, Robert Giaquinto, Haifeng Qian, Murali~Krishna Ramanathan, Ramesh Nallapati, Baishakhi Ray, Parminder Bhatia, Sudipta Sengupta, Dan Roth, and Bing Xiang. 2023.
\newblock \href {https://openreview.net/forum?id=Bo7eeXm6An8} {Multi-lingual evaluation of code generation models}.
\newblock In \emph{The Eleventh International Conference on Learning Representations}.

\bibitem[{Auer et~al.(2007)Auer, Bizer, Kobilarov, Lehmann, Cyganiak, and Ives}]{10.1007/978-3-540-76298-0_52}
S{\"o}ren Auer, Christian Bizer, Georgi Kobilarov, Jens Lehmann, Richard Cyganiak, and Zachary Ives. 2007.
\newblock Dbpedia: A nucleus for a web of open data.
\newblock In \emph{The Semantic Web}, pages 722--735, Berlin, Heidelberg. Springer Berlin Heidelberg.

\bibitem[{Bojar et~al.(2014)Bojar, Buck, Federmann, Haddow, Koehn, Leveling, Monz, Pecina, Post, Saint-Amand, Soricut, Specia, and Tamchyna}]{bojar-etal-2014-findings}
Ond{\v{r}}ej Bojar, Christian Buck, Christian Federmann, Barry Haddow, Philipp Koehn, Johannes Leveling, Christof Monz, Pavel Pecina, Matt Post, Herve Saint-Amand, Radu Soricut, Lucia Specia, and Ale{\v{s}} Tamchyna. 2014.
\newblock \href {https://doi.org/10.3115/v1/W14-3302} {Findings of the 2014 workshop on statistical machine translation}.
\newblock In \emph{Proceedings of the Ninth Workshop on Statistical Machine Translation}, pages 12--58, Baltimore, Maryland, USA. Association for Computational Linguistics.

\bibitem[{Brassil et~al.(1995)Brassil, Low, Maxemchuk, and O'Gorman}]{464718}
J.T. Brassil, S.~Low, N.F. Maxemchuk, and L.~O'Gorman. 1995.
\newblock \href {https://doi.org/10.1109/49.464718} {Electronic marking and identification techniques to discourage document copying}.
\newblock \emph{IEEE Journal on Selected Areas in Communications}, 13(8):1495--1504.

\bibitem[{Cettolo et~al.(2014)Cettolo, Niehues, Stuker, Bentivogli, and Federico}]{CettoloNiehuesStueker2014_1000166288}
Mauro Cettolo, Jan Niehues, Sebastian Stuker, Luisa Bentivogli, and Marcello Federico. 2014.
\newblock Report on the 11th iwslt evaluation campaign.
\newblock In \emph{Proceedings of the 11th International Workshop on Spoken Language Translation: Evaluation Campaign. Ed.: M. Federico, S. Stuker, F. Yvon}, page 2‚Äì17. {Association for Computational Linguistics (ACL)}.

\bibitem[{Chang et~al.(2024)Chang, Krishna, Houmansadr, Wieting, and Iyyer}]{chang2024postmark}
Yapei Chang, Kalpesh Krishna, Amir Houmansadr, John Wieting, and Mohit Iyyer. 2024.
\newblock Postmark: A robust blackbox watermark for large language models.
\newblock \emph{arXiv preprint arXiv:2406.14517}.

\bibitem[{Chen et~al.(2023)Chen, Bian, Deng, Li, Wu, Zhao, and Wong}]{chen2023x}
Liang Chen, Yatao Bian, Yang Deng, Shuaiyi Li, Bingzhe Wu, Peilin Zhao, and Kam-fai Wong. 2023.
\newblock X-mark: Towards lossless watermarking through lexical redundancy.
\newblock \emph{arXiv preprint arXiv:2311.09832}.

\bibitem[{Chen et~al.(2021)Chen, Tworek, Jun, Yuan, Pinto, Kaplan, Edwards, Burda, Joseph, Brockman et~al.}]{chen2021evaluating}
Mark Chen, Jerry Tworek, Heewoo Jun, Qiming Yuan, Henrique Ponde de~Oliveira Pinto, Jared Kaplan, Harri Edwards, Yuri Burda, Nicholas Joseph, Greg Brockman, et~al. 2021.
\newblock Evaluating large language models trained on code.
\newblock \emph{arXiv preprint arXiv:2107.03374}.

\bibitem[{Dai et~al.(2022)Dai, Mao, Fan, and Zhou}]{dai2022deephider}
Long Dai, Jiarong Mao, Xuefeng Fan, and Xiaoyi Zhou. 2022.
\newblock Deephider: A covert nlp watermarking framework based on multi-task learning.
\newblock \emph{arXiv preprint arXiv:2208.04676}.

\bibitem[{Fairoze et~al.(2023)Fairoze, Garg, Jha, Mahloujifar, Mahmoody, and Wang}]{cryptoeprint:2023/1661}
Jaiden Fairoze, Sanjam Garg, Somesh Jha, Saeed Mahloujifar, Mohammad Mahmoody, and Mingyuan Wang. 2023.
\newblock \href {https://eprint.iacr.org/2023/1661} {Publicly detectable watermarking for language models}.
\newblock Cryptology ePrint Archive, Paper 2023/1661.
\newblock \url{https://eprint.iacr.org/2023/1661}.

\bibitem[{Fu et~al.(2024)Fu, Xiong, and Dong}]{fu2024watermarking}
Yu~Fu, Deyi Xiong, and Yue Dong. 2024.
\newblock Watermarking conditional text generation for ai detection: Unveiling challenges and a semantic-aware watermark remedy.
\newblock In \emph{Proceedings of the AAAI Conference on Artificial Intelligence}, volume~38, pages 18003--18011.

\bibitem[{Gardent et~al.(2017)Gardent, Shimorina, Narayan, and Perez-Beltrachini}]{gardent-etal-2017-creating}
Claire Gardent, Anastasia Shimorina, Shashi Narayan, and Laura Perez-Beltrachini. 2017.
\newblock \href {https://doi.org/10.18653/v1/P17-1017} {Creating training corpora for {NLG} micro-planners}.
\newblock In \emph{Proceedings of the 55th Annual Meeting of the Association for Computational Linguistics (Volume 1: Long Papers)}, pages 179--188, Vancouver, Canada. Association for Computational Linguistics.

\bibitem[{Gerlach and Font-Clos(2020)}]{gerlach2020standardized}
Martin Gerlach and Francesc Font-Clos. 2020.
\newblock A standardized project gutenberg corpus for statistical analysis of natural language and quantitative linguistics.
\newblock \emph{Entropy}, 22(1):126.

\bibitem[{Guan et~al.(2024)Guan, Wan, Bi, Wang, Zhang, Sui, Zhou, and Sun}]{guan2024codeip}
Batu Guan, Yao Wan, Zhangqian Bi, Zheng Wang, Hongyu Zhang, Yulei Sui, Pan Zhou, and Lichao Sun. 2024.
\newblock Codeip: A grammar-guided multi-bit watermark for large language models of code.
\newblock \emph{arXiv preprint arXiv:2404.15639}.

\bibitem[{Guo et~al.(2023)Guo, Zhang, Wang, Jiang, Nie, Ding, Yue, and Wu}]{guo2023close}
Biyang Guo, Xin Zhang, Ziyuan Wang, Minqi Jiang, Jinran Nie, Yuxuan Ding, Jianwei Yue, and Yupeng Wu. 2023.
\newblock How close is chatgpt to human experts? comparison corpus, evaluation, and detection.
\newblock \emph{arXiv preprint arXiv:2301.07597}.

\bibitem[{He et~al.(2022{\natexlab{a}})He, Xu, Lyu, Wu, and Wang}]{he2022protecting}
Xuanli He, Qiongkai Xu, Lingjuan Lyu, Fangzhao Wu, and Chenguang Wang. 2022{\natexlab{a}}.
\newblock Protecting intellectual property of language generation apis with lexical watermark.
\newblock In \emph{Proceedings of the AAAI Conference on Artificial Intelligence}, volume~36, pages 10758--10766.

\bibitem[{He et~al.(2022{\natexlab{b}})He, Xu, Zeng, Lyu, Wu, Li, and Jia}]{he2022cater}
Xuanli He, Qiongkai Xu, Yi~Zeng, Lingjuan Lyu, Fangzhao Wu, Jiwei Li, and Ruoxi Jia. 2022{\natexlab{b}}.
\newblock Cater: Intellectual property protection on text generation apis via conditional watermarks.
\newblock \emph{Advances in Neural Information Processing Systems}, 35:5431--5445.

\bibitem[{Hendrycks et~al.(2021)Hendrycks, Basart, Kadavath, Mazeika, Arora, Guo, Burns, Puranik, He, Song et~al.}]{hendrycks2021measuring}
Dan Hendrycks, Steven Basart, Saurav Kadavath, Mantas Mazeika, Akul Arora, Ethan Guo, Collin Burns, Samir Puranik, Horace He, Dawn Song, et~al. 2021.
\newblock Measuring coding challenge competence with apps.
\newblock \emph{arXiv preprint arXiv:2105.09938}.

\bibitem[{Hoang et~al.(2024)Hoang, Le, Chu, Li, Zhao, Lao, and Doan}]{hoang2024less}
Duy~C Hoang, Hung~TQ Le, Rui Chu, Ping Li, Weijie Zhao, Yingjie Lao, and Khoa~D Doan. 2024.
\newblock Less is more: Sparse watermarking in llms with enhanced text quality.
\newblock \emph{arXiv preprint arXiv:2407.13803}.

\bibitem[{Hou et~al.(2023)Hou, Zhang, He, Wang, Chuang, Wang, Shen, Durme, Khashabi, and Tsvetkov}]{hou2023semstamp}
Abe~Bohan Hou, Jingyu Zhang, Tianxing He, Yichen Wang, Yung-Sung Chuang, Hongwei Wang, Lingfeng Shen, Benjamin~Van Durme, Daniel Khashabi, and Yulia Tsvetkov. 2023.
\newblock \href {https://arxiv.org/abs/2310.03991} {Semstamp: A semantic watermark with paraphrastic robustness for text generation}.
\newblock \emph{Preprint}, arXiv:2310.03991.

\bibitem[{Hu et~al.(2023)Hu, Chen, Wu, Wu, Zhang, and Huang}]{hu2023unbiased}
Zhengmian Hu, Lichang Chen, Xidong Wu, Yihan Wu, Hongyang Zhang, and Heng Huang. 2023.
\newblock Unbiased watermark for large language models.
\newblock \emph{arXiv preprint arXiv:2310.10669}.

\bibitem[{Husain et~al.(2019)Husain, Wu, Gazit, Allamanis, and Brockschmidt}]{husain2019codesearchnet}
Hamel Husain, Ho-Hsiang Wu, Tiferet Gazit, Miltiadis Allamanis, and Marc Brockschmidt. 2019.
\newblock Codesearchnet challenge: Evaluating the state of semantic code search.
\newblock \emph{arXiv preprint arXiv:1909.09436}.

\bibitem[{Jalil and Mirza(2009)}]{jalil2009review}
Zunera Jalil and Anwar~M Mirza. 2009.
\newblock A review of digital watermarking techniques for text documents.
\newblock In \emph{2009 International Conference on Information and Multimedia Technology}, pages 230--234. IEEE.

\bibitem[{Kamaruddin et~al.(2018)Kamaruddin, Kamsin, Por, and Rahman}]{kamaruddin2018review}
Nurul~Shamimi Kamaruddin, Amirrudin Kamsin, Lip~Yee Por, and Hameedur Rahman. 2018.
\newblock A review of text watermarking: theory, methods, and applications.
\newblock \emph{IEEE Access}, 6:8011--8028.

\bibitem[{Kirchenbauer et~al.(2023)Kirchenbauer, Geiping, Wen, Katz, Miers, and Goldstein}]{kirchenbauer2023watermark}
John Kirchenbauer, Jonas Geiping, Yuxin Wen, Jonathan Katz, Ian Miers, and Tom Goldstein. 2023.
\newblock A watermark for large language models.
\newblock In \emph{International Conference on Machine Learning}, pages 17061--17084. PMLR.

\bibitem[{Krishna et~al.(2024)Krishna, Song, Karpinska, Wieting, and Iyyer}]{krishna2024paraphrasing}
Kalpesh Krishna, Yixiao Song, Marzena Karpinska, John Wieting, and Mohit Iyyer. 2024.
\newblock Paraphrasing evades detectors of ai-generated text, but retrieval is an effective defense.
\newblock \emph{Advances in Neural Information Processing Systems}, 36.

\bibitem[{Kuditipudi et~al.(2023)Kuditipudi, Thickstun, Hashimoto, and Liang}]{kuditipudi2023robust}
Rohith Kuditipudi, John Thickstun, Tatsunori Hashimoto, and Percy Liang. 2023.
\newblock Robust distortion-free watermarks for language models.
\newblock \emph{arXiv preprint arXiv:2307.15593}.

\bibitem[{Kumar et~al.(2024)Kumar, Singh, Murty, and Ragupathy}]{kumar2024ethics}
Ashutosh Kumar, Sagarika Singh, Shiv~Vignesh Murty, and Swathy Ragupathy. 2024.
\newblock The ethics of interaction: Mitigating security threats in llms.
\newblock \emph{arXiv preprint arXiv:2401.12273}.

\bibitem[{Lai et~al.(2023)Lai, Li, Wang, Zhang, Zhong, Zettlemoyer, Yih, Fried, Wang, and Yu}]{lai2023ds}
Yuhang Lai, Chengxi Li, Yiming Wang, Tianyi Zhang, Ruiqi Zhong, Luke Zettlemoyer, Wen-tau Yih, Daniel Fried, Sida Wang, and Tao Yu. 2023.
\newblock Ds-1000: A natural and reliable benchmark for data science code generation.
\newblock In \emph{International Conference on Machine Learning}, pages 18319--18345. PMLR.

\bibitem[{Lee et~al.(2023)Lee, Hong, Ahn, Hong, Lee, Yun, Shin, and Kim}]{lee2023wrote}
Taehyun Lee, Seokhee Hong, Jaewoo Ahn, Ilgee Hong, Hwaran Lee, Sangdoo Yun, Jamin Shin, and Gunhee Kim. 2023.
\newblock Who wrote this code? watermarking for code generation.
\newblock \emph{arXiv preprint arXiv:2305.15060}.

\bibitem[{Lewis et~al.(2004)Lewis, Yang, Russell-Rose, and Li}]{lewis2004rcv1}
David~D Lewis, Yiming Yang, Tony Russell-Rose, and Fan Li. 2004.
\newblock Rcv1: A new benchmark collection for text categorization research.
\newblock \emph{Journal of machine learning research}, 5(Apr):361--397.

\bibitem[{Li et~al.(2024)Li, Zhang, Zhang, Sun, and Wang}]{li2024resilient}
Boquan Li, Mengdi Zhang, Peixin Zhang, Jun Sun, and Xingmei Wang. 2024.
\newblock Resilient watermarking for llm-generated codes.
\newblock \emph{arXiv preprint arXiv:2402.07518}.

\bibitem[{Li et~al.(2023)Li, Yang, Sun, Chen, Song, Xiang, Wang, and Zhou}]{li2023towards}
Wei Li, Borui Yang, Yujie Sun, Suyu Chen, Ziyun Song, Liyao Xiang, Xinbing Wang, and Chenghu Zhou. 2023.
\newblock Towards tracing code provenance with code watermarking.
\newblock \emph{arXiv preprint arXiv:2305.12461}.

\bibitem[{Lin et~al.(2021)Lin, Hilton, and Evans}]{lin2021truthfulqa}
Stephanie Lin, Jacob Hilton, and Owain Evans. 2021.
\newblock Truthfulqa: Measuring how models mimic human falsehoods.
\newblock \emph{arXiv preprint arXiv:2109.07958}.

\bibitem[{Liu et~al.(2023{\natexlab{a}})Liu, Pan, Hu, Li, Wen, King, and Yu}]{liu2023private}
Aiwei Liu, Leyi Pan, Xuming Hu, Shu'ang Li, Lijie Wen, Irwin King, and Philip~S Yu. 2023{\natexlab{a}}.
\newblock A private watermark for large language models.
\newblock \emph{arXiv preprint arXiv:2307.16230}.

\bibitem[{Liu et~al.(2023{\natexlab{b}})Liu, Pan, Hu, Meng, and Wen}]{liu2023semantic}
Aiwei Liu, Leyi Pan, Xuming Hu, Shiao Meng, and Lijie Wen. 2023{\natexlab{b}}.
\newblock A semantic invariant robust watermark for large language models.
\newblock \emph{arXiv preprint arXiv:2310.06356}.

\bibitem[{Liu and Bu(2024)}]{liu2024adaptive}
Yepeng Liu and Yuheng Bu. 2024.
\newblock Adaptive text watermark for large language models.
\newblock \emph{arXiv preprint arXiv:2401.13927}.

\bibitem[{Liu et~al.(2023{\natexlab{c}})Liu, Hu, Zhang, and Sun}]{liu2023watermarking}
Yixin Liu, Hongsheng Hu, Xuyun Zhang, and Lichao Sun. 2023{\natexlab{c}}.
\newblock Watermarking text data on large language models for dataset copyright protection.
\newblock \emph{arXiv preprint arXiv:2305.13257}.

\bibitem[{Maas et~al.(2011)Maas, Daly, Pham, Huang, Ng, and Potts}]{maas-etal-2011-learning}
Andrew~L. Maas, Raymond~E. Daly, Peter~T. Pham, Dan Huang, Andrew~Y. Ng, and Christopher Potts. 2011.
\newblock \href {https://aclanthology.org/P11-1015} {Learning word vectors for sentiment analysis}.
\newblock In \emph{Proceedings of the 49th Annual Meeting of the Association for Computational Linguistics: Human Language Technologies}, pages 142--150, Portland, Oregon, USA. Association for Computational Linguistics.

\bibitem[{Magnusson et~al.(2023)Magnusson, Bhagia, Hofmann, Soldaini, Jha, Tafjord, Schwenk, Walsh, Elazar, Lo et~al.}]{magnusson2023paloma}
Ian Magnusson, Akshita Bhagia, Valentin Hofmann, Luca Soldaini, Ananya~Harsh Jha, Oyvind Tafjord, Dustin Schwenk, Evan~Pete Walsh, Yanai Elazar, Kyle Lo, et~al. 2023.
\newblock Paloma: A benchmark for evaluating language model fit.
\newblock \emph{arXiv preprint arXiv:2312.10523}.

\bibitem[{Mao et~al.(2024)Mao, Wei, Chen, Fang, and Chau}]{mao2024watermark}
Minjia Mao, Dongjun Wei, Zeyu Chen, Xiao Fang, and Michael Chau. 2024.
\newblock A watermark for low-entropy and unbiased generation in large language models.
\newblock \emph{arXiv preprint arXiv:2405.14604}.

\bibitem[{Meral et~al.(2009)Meral, Sankur, {\"O}zsoy, G{\"u}ng{\"o}r, and Sevin{\c{c}}}]{meral2009natural}
Hasan~Mesut Meral, B{\"u}lent Sankur, A~Sumru {\"O}zsoy, Tunga G{\"u}ng{\"o}r, and Emre Sevin{\c{c}}. 2009.
\newblock Natural language watermarking via morphosyntactic alterations.
\newblock \emph{Computer Speech \& Language}, 23(1):107--125.

\bibitem[{Merity et~al.(2016)Merity, Xiong, Bradbury, and Socher}]{merity2016pointer}
Stephen Merity, Caiming Xiong, James Bradbury, and Richard Socher. 2016.
\newblock Pointer sentinel mixture models.
\newblock \emph{arXiv preprint arXiv:1609.07843}.

\bibitem[{Metsis et~al.(2006)Metsis, Androutsopoulos, and Paliouras}]{inproceedings}
Vangelis Metsis, Ion Androutsopoulos, and Georgios Paliouras. 2006.
\newblock Spam filtering with naive bayes - which naive bayes?

\bibitem[{Mostafazadeh et~al.(2016)Mostafazadeh, Chambers, He, Parikh, Batra, Vanderwende, Kohli, and Allen}]{mostafazadeh2016corpus}
Nasrin Mostafazadeh, Nathanael Chambers, Xiaodong He, Devi Parikh, Dhruv Batra, Lucy Vanderwende, Pushmeet Kohli, and James Allen. 2016.
\newblock A corpus and evaluation framework for deeper understanding of commonsense stories.
\newblock \emph{arXiv preprint arXiv:1604.01696}.

\bibitem[{Munyer and Zhong(2023)}]{munyer2023deeptextmark}
Travis Munyer and Xin Zhong. 2023.
\newblock Deeptextmark: Deep learning based text watermarking for detection of large language model generated text.
\newblock \emph{arXiv preprint arXiv:2305.05773}.

\bibitem[{Nallapati et~al.(2016)Nallapati, Zhou, Gulcehre, Xiang et~al.}]{nallapati2016abstractive}
Ramesh Nallapati, Bowen Zhou, Caglar Gulcehre, Bing Xiang, et~al. 2016.
\newblock Abstractive text summarization using sequence-to-sequence rnns and beyond.
\newblock \emph{arXiv preprint arXiv:1602.06023}.

\bibitem[{Nan et~al.(2021)Nan, Radev, Zhang, Rau, Sivaprasad, Hsieh, Tang, Vyas, Verma, Krishna, Liu, Irwanto, Pan, Rahman, Zaidi, Mutuma, Tarabar, Gupta, Yu, Tan, Lin, Xiong, Socher, and Rajani}]{nan-etal-2021-dart}
Linyong Nan, Dragomir Radev, Rui Zhang, Amrit Rau, Abhinand Sivaprasad, Chiachun Hsieh, Xiangru Tang, Aadit Vyas, Neha Verma, Pranav Krishna, Yangxiaokang Liu, Nadia Irwanto, Jessica Pan, Faiaz Rahman, Ahmad Zaidi, Mutethia Mutuma, Yasin Tarabar, Ankit Gupta, Tao Yu, Yi~Chern Tan, Xi~Victoria Lin, Caiming Xiong, Richard Socher, and Nazneen~Fatema Rajani. 2021.
\newblock \href {https://doi.org/10.18653/v1/2021.naacl-main.37} {{DART}: Open-domain structured data record to text generation}.
\newblock In \emph{Proceedings of the 2021 Conference of the North American Chapter of the Association for Computational Linguistics: Human Language Technologies}, pages 432--447, Online. Association for Computational Linguistics.

\bibitem[{Narayan et~al.(2018)Narayan, Cohen, and Lapata}]{narayan2018don}
Shashi Narayan, Shay~B Cohen, and Mirella Lapata. 2018.
\newblock Don't give me the details, just the summary! topic-aware convolutional neural networks for extreme summarization.
\newblock \emph{arXiv preprint arXiv:1808.08745}.

\bibitem[{{New York Times Company}(2023)}]{initial_suit}
The {New York Times Company}. 2023.
\newblock \href {https://nytco-assets.nytimes.com/2023/12/NYT_Complaint_Dec2023.pdf} {The new york times company v. microsoft corporation, openai, inc., openai lp, openai gp, llc, openai, llc, openai opco llc, openai global llc, oai corporation, llc, and openai holdings, llc}.

\bibitem[{{Nicolai Thorer Sivesind}(2023)}]{sivesind_2023}
{Nicolai Thorer Sivesind}. 2023.
\newblock Chatgpt-generated-abstracts.

\bibitem[{Peng et~al.(2023)Peng, Yi, Wu, Wu, Zhu, Lyu, Jiao, Xu, Sun, and Xie}]{peng2023you}
Wenjun Peng, Jingwei Yi, Fangzhao Wu, Shangxi Wu, Bin Zhu, Lingjuan Lyu, Binxing Jiao, Tong Xu, Guangzhong Sun, and Xing Xie. 2023.
\newblock Are you copying my model? protecting the copyright of large language models for eaas via backdoor watermark.
\newblock \emph{arXiv preprint arXiv:2305.10036}.

\bibitem[{Por et~al.(2012)Por, Wong, and Chee}]{POR20121075}
Lip~Yee Por, KokSheik Wong, and Kok~Onn Chee. 2012.
\newblock \href {https://doi.org/10.1016/j.jss.2011.12.023} {Unispach: A text-based data hiding method using unicode space characters}.
\newblock \emph{Journal of Systems and Software}, 85(5):1075--1082.

\bibitem[{Qu et~al.(2024)Qu, Yin, He, Zou, Tao, Jia, and Zhang}]{qu2024provably}
Wenjie Qu, Dong Yin, Zixin He, Wei Zou, Tianyang Tao, Jinyuan Jia, and Jiaheng Zhang. 2024.
\newblock Provably robust multi-bit watermarking for ai-generated text via error correction code.
\newblock \emph{arXiv preprint arXiv:2401.16820}.

\bibitem[{Raffel et~al.(2020)Raffel, Shazeer, Roberts, Lee, Narang, Matena, Zhou, Li, and Liu}]{raffel2020exploring}
Colin Raffel, Noam Shazeer, Adam Roberts, Katherine Lee, Sharan Narang, Michael Matena, Yanqi Zhou, Wei Li, and Peter~J Liu. 2020.
\newblock Exploring the limits of transfer learning with a unified text-to-text transformer.
\newblock \emph{Journal of machine learning research}, 21(140):1--67.

\bibitem[{Ren et~al.(2023)Ren, Xu, Liu, Cui, Wang, Yin, and Tang}]{ren2023robust}
Jie Ren, Han Xu, Yiding Liu, Yingqian Cui, Shuaiqiang Wang, Dawei Yin, and Jiliang Tang. 2023.
\newblock A robust semantics-based watermark for large language model against paraphrasing.
\newblock \emph{arXiv preprint arXiv:2311.08721}.

\bibitem[{Ren et~al.(2024)Ren, Guo, Cao, and Ma}]{ren-etal-2024-subtle}
Yubing Ren, Ping Guo, Yanan Cao, and Wei Ma. 2024.
\newblock \href {https://doi.org/10.18653/v1/2024.findings-acl.327} {Subtle signatures, strong shields: Advancing robust and imperceptible watermarking in large language models}.
\newblock In \emph{Findings of the Association for Computational Linguistics: ACL 2024}, pages 5508--5519, Bangkok, Thailand. Association for Computational Linguistics.

\bibitem[{Sato et~al.(2023)Sato, Takezawa, Bao, Niwa, and Yamada}]{sato2023embarrassingly}
Ryoma Sato, Yuki Takezawa, Han Bao, Kenta Niwa, and Makoto Yamada. 2023.
\newblock Embarrassingly simple text watermarks.
\newblock \emph{arXiv preprint arXiv:2310.08920}.

\bibitem[{Shao et~al.(2024)Shao, Li, Yao, He, Qin, and Ren}]{shao2024explanation}
Shuo Shao, Yiming Li, Hongwei Yao, Yiling He, Zhan Qin, and Kui Ren. 2024.
\newblock Explanation as a watermark: Towards harmless and multi-bit model ownership verification via watermarking feature attribution.
\newblock \emph{arXiv preprint arXiv:2405.04825}.

\bibitem[{Socher et~al.(2013)Socher, Perelygin, Wu, Chuang, Manning, Ng, and Potts}]{socher-etal-2013-recursive}
Richard Socher, Alex Perelygin, Jean Wu, Jason Chuang, Christopher~D. Manning, Andrew Ng, and Christopher Potts. 2013.
\newblock \href {https://aclanthology.org/D13-1170} {Recursive deep models for semantic compositionality over a sentiment treebank}.
\newblock In \emph{Proceedings of the 2013 Conference on Empirical Methods in Natural Language Processing}, pages 1631--1642, Seattle, Washington, USA. Association for Computational Linguistics.

\bibitem[{Sun et~al.(2023)Sun, Du, Song, and Li}]{sun2023codemark}
Zhensu Sun, Xiaoning Du, Fu~Song, and Li~Li. 2023.
\newblock Codemark: Imperceptible watermarking for code datasets against neural code completion models.
\newblock In \emph{Proceedings of the 31st ACM Joint European Software Engineering Conference and Symposium on the Foundations of Software Engineering}, pages 1561--1572.

\bibitem[{Takezawa et~al.(2023)Takezawa, Sato, Bao, Niwa, and Yamada}]{takezawa2023necessary}
Yuki Takezawa, Ryoma Sato, Han Bao, Kenta Niwa, and Makoto Yamada. 2023.
\newblock Necessary and sufficient watermark for large language models.
\newblock \emph{arXiv preprint arXiv:2310.00833}.

\bibitem[{Tang et~al.(2023)Tang, Feng, Liu, Yang, and Hu}]{tang2023did}
Ruixiang Tang, Qizhang Feng, Ninghao Liu, Fan Yang, and Xia Hu. 2023.
\newblock Did you train on my dataset? towards public dataset protection with cleanlabel backdoor watermarking.
\newblock \emph{ACM SIGKDD Explorations Newsletter}, 25(1):43--53.

\bibitem[{Topkara et~al.(2006{\natexlab{a}})Topkara, Topkara, and Atallah}]{Topkara2006WordsAN}
Mercan Topkara, Umut Topkara, and Mikhail~J. Atallah. 2006{\natexlab{a}}.
\newblock \href {https://api.semanticscholar.org/CorpusID:5854860} {Words are not enough: sentence level natural language watermarking}.
\newblock In \emph{Workshop on Medical Cyber-Physical Systems}.

\bibitem[{Topkara et~al.(2006{\natexlab{b}})Topkara, Topkara, and Atallah}]{10.1145/1161366.1161397}
Umut Topkara, Mercan Topkara, and Mikhail~J. Atallah. 2006{\natexlab{b}}.
\newblock \href {https://doi.org/10.1145/1161366.1161397} {The hiding virtues of ambiguity: quantifiably resilient watermarking of natural language text through synonym substitutions}.
\newblock In \emph{Proceedings of the 8th Workshop on Multimedia and Security}, MM and Sec '06, page 164–174, New York, NY, USA. Association for Computing Machinery.

\bibitem[{Wang et~al.(2023)Wang, Yang, Chen, Zhou, Lin, Meng, Zhou, and Sun}]{wang2023towards}
Lean Wang, Wenkai Yang, Deli Chen, Hao Zhou, Yankai Lin, Fandong Meng, Jie Zhou, and Xu~Sun. 2023.
\newblock Towards codable text watermarking for large language models.
\newblock \emph{arXiv preprint arXiv:2307.15992}.

\bibitem[{Wu et~al.(2020)Wu, Qiao, Chen, Wu, Qi, Lian, Liu, Xie, Gao, Wu, and Zhou}]{wu-etal-2020-mind}
Fangzhao Wu, Ying Qiao, Jiun-Hung Chen, Chuhan Wu, Tao Qi, Jianxun Lian, Danyang Liu, Xing Xie, Jianfeng Gao, Winnie Wu, and Ming Zhou. 2020.
\newblock \href {https://doi.org/10.18653/v1/2020.acl-main.331} {{MIND}: A large-scale dataset for news recommendation}.
\newblock In \emph{Proceedings of the 58th Annual Meeting of the Association for Computational Linguistics}, pages 3597--3606, Online. Association for Computational Linguistics.

\bibitem[{Wu et~al.(2023)Wu, Hu, Zhang, and Huang}]{wu2023dipmark}
Yihan Wu, Zhengmian Hu, Hongyang Zhang, and Heng Huang. 2023.
\newblock Dipmark: A stealthy, efficient and resilient watermark for large language models.
\newblock \emph{arXiv preprint arXiv:2310.07710}.

\bibitem[{Yang et~al.(2023{\natexlab{a}})Yang, Li, Xiang, and Li}]{yang2023towards}
Borui Yang, Wei Li, Liyao Xiang, and Bo~Li. 2023{\natexlab{a}}.
\newblock Towards code watermarking with dual-channel transformations.
\newblock \emph{arXiv preprint arXiv:2309.00860}.

\bibitem[{Yang et~al.(2023{\natexlab{b}})Yang, Chen, Zhang, Liu, Qi, Zhang, Fang, and Yu}]{yang2023watermarking}
Xi~Yang, Kejiang Chen, Weiming Zhang, Chang Liu, Yuang Qi, Jie Zhang, Han Fang, and Nenghai Yu. 2023{\natexlab{b}}.
\newblock Watermarking text generated by black-box language models.
\newblock \emph{arXiv preprint arXiv:2305.08883}.

\bibitem[{Yang et~al.(2022)Yang, Zhang, Chen, Zhang, Ma, Wang, and Yu}]{yang2022tracing}
Xi~Yang, Jie Zhang, Kejiang Chen, Weiming Zhang, Zehua Ma, Feng Wang, and Nenghai Yu. 2022.
\newblock Tracing text provenance via context-aware lexical substitution.
\newblock In \emph{Proceedings of the AAAI Conference on Artificial Intelligence}, volume~36, pages 11613--11621.

\bibitem[{Yoo et~al.(2023{\natexlab{a}})Yoo, Ahn, Jang, and Kwak}]{yoo2023robust}
KiYoon Yoo, Wonhyuk Ahn, Jiho Jang, and Nojun Kwak. 2023{\natexlab{a}}.
\newblock Robust natural language watermarking through invariant features.
\newblock \emph{arXiv preprint arXiv:2305.01904}.

\bibitem[{Yoo et~al.(2023{\natexlab{b}})Yoo, Ahn, and Kwak}]{yoo2023advancing}
KiYoon Yoo, Wonhyuk Ahn, and Nojun Kwak. 2023{\natexlab{b}}.
\newblock Advancing beyond identification: Multi-bit watermark for language models.
\newblock \emph{arXiv preprint arXiv:2308.00221}.

\bibitem[{Zhang et~al.(2023)Zhang, Hussain, Neekhara, and Koushanfar}]{zhang2023remark}
Ruisi Zhang, Shehzeen~Samarah Hussain, Paarth Neekhara, and Farinaz Koushanfar. 2023.
\newblock Remark-llm: A robust and efficient watermarking framework for generative large language models.
\newblock \emph{arXiv preprint arXiv:2310.12362}.

\bibitem[{Zhang et~al.(2015)Zhang, Zhao, and LeCun}]{zhang2015character}
Xiang Zhang, Junbo Zhao, and Yann LeCun. 2015.
\newblock Character-level convolutional networks for text classification.
\newblock \emph{Advances in neural information processing systems}, 28.

\bibitem[{Zhao et~al.(2023{\natexlab{a}})Zhao, Ananth, Li, and Wang}]{zhao2023provable}
Xuandong Zhao, Prabhanjan Ananth, Lei Li, and Yu-Xiang Wang. 2023{\natexlab{a}}.
\newblock Provable robust watermarking for ai-generated text.
\newblock \emph{arXiv preprint arXiv:2306.17439}.

\bibitem[{Zhao et~al.(2023{\natexlab{b}})Zhao, Wang, and Li}]{zhao2023protecting}
Xuandong Zhao, Yu-Xiang Wang, and Lei Li. 2023{\natexlab{b}}.
\newblock Protecting language generation models via invisible watermarking.
\newblock In \emph{International Conference on Machine Learning}, pages 42187--42199. PMLR.

\bibitem[{Zheng et~al.(2024)Zheng, Chiang, Sheng, Zhuang, Wu, Zhuang, Lin, Li, Li, Xing et~al.}]{zheng2024judging}
Lianmin Zheng, Wei-Lin Chiang, Ying Sheng, Siyuan Zhuang, Zhanghao Wu, Yonghao Zhuang, Zi~Lin, Zhuohan Li, Dacheng Li, Eric Xing, et~al. 2024.
\newblock Judging llm-as-a-judge with mt-bench and chatbot arena.
\newblock \emph{Advances in Neural Information Processing Systems}, 36.

\bibitem[{Zhou et~al.(2024)Zhou, Zhao, Xu, and Ren}]{zhou2024bileve}
Tong Zhou, Xuandong Zhao, Xiaolin Xu, and Shaolei Ren. 2024.
\newblock Bileve: Securing text provenance in large language models against spoofing with bi-level signature.
\newblock \emph{arXiv preprint arXiv:2406.01946}.

\end{thebibliography}

\section{Appendix}









\subsection{Watermarking and Steganography} \label{stegano}

The concepts of text watermarking and steganography are often discussed together due to their shared goal of altering text to convey additional information. However, they serve distinct purposes and have different applications.

\textbf{Steganography} is the practice of concealing messages within non-secret text or data, making the hidden message indiscernible to unintended recipients. The primary objective of steganography is to ensure that the existence of the concealed message remains undetected. This is often achieved through subtle alterations to the host text that are imperceptible to the human eye or standard detection techniques. Steganography is widely used in fields such as secure communications and digital rights management.

\textbf{Text Watermarking}, on the other hand, involves embedding identifiable information or markers within text to establish ownership, verify authenticity, or detect unauthorized use. Unlike steganography, text watermarking does not necessarily seek to conceal the existence of the watermark but rather focuses on making it robust and detectable under various conditions. The key objectives of text watermarking include ensuring that the watermarked content remains readable and maintaining the natural quality of the text, while providing a reliable means of verification or ownership assertion.

While both steganography and text watermarking involve altering text, they differ significantly in their underlying intentions and techniques. Steganography prioritizes secrecy and concealment, whereas text watermarking emphasizes detection, ownership, and authenticity. Understanding these differences is crucial when selecting the appropriate technique for specific applications.

\subsection{Some metrics used for evaluation} \label{metrics for evaluation}

To evaluate the effectiveness of watermarking techniques, we consider two categories of metrics: \textbf{detection metrics} and \textbf{downstream task evaluation metrics}.

For watermark detection, we evaluate the following metrics:

\textbf{1. Watermark Success Rate:} The percentage of cases where the embedded watermark is successfully detected, indicating the reliability of the watermarking technique.

\textbf{2. Area Under the Curve (AUC):} Represents the model's ability to distinguish between watermarked and non-watermarked texts, derived from the ROC (Receiver Operating Characteristic) curve. A higher AUC indicates better performance in correctly classifying watermarked and non-watermarked text.

\textbf{3. False Positive Rate (FPR):} Measures the rate at which non-watermarked texts are incorrectly identified as watermarked, which is crucial for assessing the precision of watermark detection methods.

\begin{equation}
\text{FPR} = \frac{\text{False Positives}}{\text{False Positives} + \text{True Negatives}}
\end{equation}

To ensure that watermarking does not adversely affect the utility of text in various NLP tasks, we evaluate the following metrics:

\textbf{1. F1 Score:} Used for classification tasks, such as sentiment analysis or spam detection. It is the harmonic mean of precision and recall, evaluating the balance between false positives and false negatives.

\begin{equation}
\text{F1 Score} = 2 \cdot \frac{\text{Precision} \cdot \text{Recall}}{\text{Precision} + \text{Recall}}
\end{equation}

\textbf{2. Exact Match (EM):} Used primarily in question answering tasks, this metric measures the percentage of predictions that exactly match the ground truth answers.

\[
\text{EM} = \frac{\text{Number of Exact Matches}}{\text{Total Number of Predictions}}
\]

\textbf{3. Perplexity (PPL):} Commonly used in language modeling and text generation tasks, perplexity measures the fluency of generated text. Lower perplexity indicates that the text is more coherent and closer to natural language usage.

To mathematically define perplexity for a sequence of words $W = (w_1, w_2, \ldots, w_N)$, it can be expressed as:
\begin{equation}
PP(W) = P(w_1 w_2 \ldots w_N)^{-\frac{1}{N}}
\end{equation}

where $P(w_1 w_2 \ldots w_N)$ is the probability of the word sequence $W$ according to the model.













\end{document}